# A New K-means Grey Wolf Algorithm for Engineering Problems

Hardi M. Mohammed[1,2], Zrar Kh. Abdul[1], Tarik A. Rashid[3,*], Abeer Alsadoon[4,5], Nebojsa Bacanin[6]

[1]Applied Computer Department, College of Medicals and Applied Sciences, Charmo University, Sulaimani, Chamchamal, KRG, Iraq.
[2]Technical College of Informatics, Sulaimani Polytechnic University, Sulaimani, KRG, Iraq,
[3]Computer Science and Engineering, University of Kurdistan Hewler, Erbil, KRG, Iraq. tarik.ahmed@ukh.edu.krd
[4]Charles Sturt University, Sydney, Australia.
[5]Information Technology Department, Asia Pacific International College (APIC), Sydney, Australia
[6]Singidunum University, Belgrade, Serbia.

## Abstract:

**Purpose:**

**This paper aims at studying meta-heuristic algorithms. One of the common meta-heuristic optimization algorithms is called grey wolf optimization (GWO). The key aim is to enhance the limitations of the wolves' searching process of attacking gray wolves.**

**Design/methodology/approach:**

**The development of meta-heuristic algorithms has increased by researchers to use them extensively in the field of business, science, and engineering. In this paper, the K-means clustering algorithm is used to enhance the performance of the original GWO; the new algorithm is called K-means clustering gray wolf optimization (KMGWO).**

**Findings:**

**Results illustrate the efficiency of KMGWO against the GWO. To evaluate the performance of the KMGWO, KMGWO was applied to solve CEC2019 benchmark test functions.**

**Originality/value:**

**Results prove that KMGWO is superior to GWO. KMGWO is also compared to cat swarm optimization (CSO), whale optimization algorithm-bat algorithm (WOA-BAT), WOA, and GWO so KMGWO achieved the first rank in terms of performance. Also, the KMGWO is used to solve a classical engineering problem and it is superior.**

# 1. Introduction

Meta-heuristic algorithms have become a common mechanism to solve and confront numerous problems by scientists and scholars. They are used specifically for those problems that have many constraints or do not have analytical solutions. Most of these algorithms were developed from the existence of the fittest evolutionary algorithms theory, the behavior of biologically inspired algorithms, and the collective of swarm intelligence particles in nature[1]. Many difficulties arise to discover a universal effective algorithm for solving nearly all problems as there is no free lunch rule. Therefore, researchers around the world keep digging to find more optimization algorithms[2].

These algorithms are used to solve real-world applications [3], [4]. These applications are divided into several areas, such as structural optimization [5], [6], [7], [8], [9], [10], scheduling and routing[11], software testing [12], image processing [13] and data mining [14], [15]. These areas of problems have been solved by a human in different ways either heuristic or metaheuristic ways. Therefore, the first use of the heuristic method was done by Alan Turing to break Enigma Cipher and then he published his invention in 1948. After that between the 1960s and 1970s is the golden period of developing an evolutionary algorithm so John Holland developed a Genetic Algorithm (GA) [16] [17], [18]. In the same decade, another search technique was developed to solve the optimization problem and called evolutionary strategy. Then, metaheuristic algorithms for example Simulated Annealing (SA) [19] and also Tabu Search [20] developed in the 1980s. Next, Ant Colony Optimization (ACO) [21], Particle Swarm Optimization (PSO) [22] and Differential Evolution (DE) [23] were developed in 1990s respectively. Harmony Search algorithm (HS) [24], Artificial Bee Colony (ABC) [25], Cuckoo Search (CS) [7] algorithm, and Bat algorithm [26] were developed in 2000 to 2010. Consequently, WOA [27], [28], GWO [29], and the most recent algorithm such as Fitness Dependent Optimizer (FDO) [30], Donkey and smuggler optimization algorithm [31], and learner performance-based behavior algorithm (LPB) [32].

were developed.

One of the popular meta-heuristic algorithms is known as Grey Wolf Optimizer, which was developed by Mirjalili et al. This algorithm is inspired by the hierarchical leadership technique and grey wolves hunting mechanism in nature [33], [34]. Implementing the GWO algorithm has been encouraging enough to merit further investigation as the GWO algorithm is used straight forward and can converge rapidly. Therefore, scientists and engineers, who are working in this field, are still under ways to improve the GWO further[1], [2], [35], [36].

Despite achieving high performance compared to other algorithms, GWO is struggling to have some issues [37]. Balancing between exploration and exploitation is one of the issues that are related to GWO [38][1]. Despite having the efficient capability, GWO has poor performance in the search space because sometimes it sticks in local optima[39], [40]. Slow convergence rates and low accuracy are problems that can be enhanced by researchers [41], [23]. Therefore, different methods are used by researchers to improve the capacity of optimization algorithms and solve optimization problems [42]–[45]. For example, a modified GWO was proposed to tune the parameters of the recurrent neural network [46]. Chaotic GWO was presented to increase the convergence speed of GWO [47]. Also, researchers have used many different technics to improve GWO [48]–[51].

Aside from swarm intelligence, extracting information from a large data set is an important concept and it is the main objective of the data mining process. The extracted information can then be transformed into an understandable structure for extra use. One of the common exploratory data analysis techniques is clustering. Clustering is used to organize objects in such a way that each cluster has more similar objects. Many cluster models have been developed to cluster the data, such as partitioning, hierarchical, density grid, and graph-based algorithms. Therefore, the partitioning clustering technique example is K-means, which works based on the centroid of the cluster[50]. K-means cluster has been used for many applications, such as in [52], [53], [54]. K-means cluster was used to image segmentation and it also was used to improve the stability of Wireless Sensor Networks [55]. Moreover, the K-means cluster has been widely used for classifying none label data as an unsupervised learning method [56], [57].

Also, K-means has been used with optimization algorithms to improve the performance of algorithms. It is simple and reliable. It also has fast convergence so that it is widely used [58]. Initializing cluster centers is one of the remarkable sensitive issues in K-means because all clusters are affected by cluster numbers. Therefore, the authors in [59] solved this problem by using GA as an optimization algorithm to optimize the number of clusters. Results showed that the proposed algorithm is better than traditional K-means. Researchers used GA to exploit K-means to maximize the performance of K-means [60]. Thus, they used K-means to divide the population into different clusters based on their similarity and then use K-means also in crossover operations as well. As a result, clustering GA proved that it is more accurate compared to K-means. In this paper [61] researchers also used GA to find the optimum number to decide which number should be used as the initial cluster number.

Because of having fast convergence capability, K-means is hybridized with PSO in [62]. therefore, PSO-KM was proposed and enhanced the capacity of PSO while it was evaluated using five datasets. PSO was used for clustering data in the beginning then if the fitness value does not change for several iterations, so, PSO clustering is switched to K-means. As a result, PSO-KM performed better than PSO and K-means clustering. K-means is also used for security evaluation in power systems [63]. K-means is nonlinear so it is not recommended to use it in a power system. In [63] K-means algorithm is modified to work as an algorithm to classify clusters. This work was done by using PSO. After checking the output of the K-means clustering by PSO results will be carried out into two different classes, which are secure class and insecure class. Results proved that PSOKM has higher accuracy and less misclassification compared to traditional K-means. Image segmentation is usually done by using K-means clustering because of its speed and simplicity [64]. However, initializing several clusters is one of the problems, which leads to falling into local optima. Therefore, its results are not satisfactory. Since PSO has strong performance in finding a global solution and low capability in falling into local optima [65]. Thus, dynamic PSOKM was proposed and results indicated that dynamic PSOKM superior to K-means clustering [64].

K-means is also used in Artificial Bee Colony (ABC) and Differential Evolution (DE). Both ABC and DE are hybridized in [66]. They used the K-means method to evaluate the fitness function for each cluster centroid. Therefore, the performance of the proposed algorithm was higher than other algorithms.

K-means is used in WOA to find the best number for clustering. So, each search agent in WOA represents the number of clusters. After that objective function is calculated by using the K-means method. Then, exploration and exploitation run. The proposed algorithm results showed high performance [67].

Proposing the KMGWO to overcome the previous issues that have been mentioned, is the main aim of this paper. The main contribution in the proposed KMGWO is to use K-means clustering after initializing the grey wolves population. Kmeans clustering is used to divide the population into two clusters and then working on each cluster separately to achieve optimality.

The rest of this paper is structured as follows; GWO is described in Section 2. Then, K-Means clustering is presented. In Section 4, KGWO is clarified. Next, investigational results are presented. After that solving pressure vessel design is represented in Section 6 and Section 7 concludes the paper.

# 2. Materials and Methods

## 2.1    Grey Wolf Optimization and K-means

The recent meta-heuristic optimization technique is the Grey wolf optimization algorithm and was first developed by [68]. It is based on the hunting behavior of grey wolves in nature to hunt prey in a supportive technique. The structure of the GWO is quite different from other meta-heuristic optimization as three optimal samples become base for a large-scale search method which are alpha (α) wolf who acts as a pack leader, then the leader is supported by beta (β) and delta (δ) is the follower of the leader and the supportive wolves. The final type of wolves is called omega (ω). These wolves are different in term of their responsibility and they can be presented as hierarchical in such a way that the top and the first solution is alpha (α), the second, third and final solutions are beta (β), delta (δ) and omega (ω) respectively. As a result, omegas are guided by the three previous types of wolves.

After finding prey, all types of wolves are trying to enclose the prey by using three different coefficient variables which are used to apply the encircling mechanism. The following equation describes the encircling method [2][69].

$$\overrightarrow{D\alpha} = \mid \overrightarrow{C_1} \cdot \overrightarrow{X\alpha} - \overrightarrow{X_i} \mid,$$
$$\overrightarrow{D_\beta} = \mid \overrightarrow{C_2} \cdot \overrightarrow{X_\beta} - \overrightarrow{X_i} \mid,$$

$$\overrightarrow{D_\delta} = \mid \overrightarrow{C_3} \cdot \overrightarrow{X_\delta} - \overrightarrow{X_i} \mid, \qquad (1)$$

Where $\overrightarrow{X}$ symbolizes the position vector of the grey wolf and $i$ is the iteration. Using three $\overrightarrow{X}(i)$ is denoted for the alpha, beta, and delta wolves. Then, each type of wolf such as alpha $\overrightarrow{X_1}$, beta $\overrightarrow{X_2}$ and delta $\overrightarrow{X_3}$ are calculated as follows in Equations 2, 3, 4, and 5:

$$\overrightarrow{X_1} = \overrightarrow{X_\alpha} - \overrightarrow{A_1} \cdot \overrightarrow{D_\alpha}, \qquad (2)$$
$$\overrightarrow{X_2} = \overrightarrow{X_\beta} - \overrightarrow{A_2} \cdot \overrightarrow{D_\beta}, \qquad (3)$$
$$\overrightarrow{X_3} = \overrightarrow{X_\delta} - \overrightarrow{A_3} \cdot \overrightarrow{D_\delta} \qquad (4)$$

$$\overrightarrow{X}(i+1) = \frac{\overrightarrow{X_1} + \overrightarrow{X_2} + \overrightarrow{X_3}}{3} \qquad (5)$$

Both *A* and *C*, are coefficient vectors, which can be calculated as follows in Equations 6 and 7:

$$\vec{A} = 2 \cdot \vec{a} \cdot \vec{r_1} + \vec{a} \qquad (6)$$
$$\vec{C} = 2 \cdot \vec{r_2} \qquad (7)$$

Variable *a* is the controlling parameter that changes the coefficient $\overrightarrow{A}$. This technique helps the omega wolf to decide to chase the prey or run away from it. Therefore, if the value of $|\vec{A}|$ is greater than 1, it means that the omega wolf tries to discover new search space. However, if the value of $|\vec{A}|$ is less than 1, the omega wolf would follow and approach the prey. Thus, this mechanism is called local search. When the hunting mechanism is achieved by doing the above producer, the grey wolf begins to avoid any movement of the prey to attack it properly. This mechanism is done by decreasing the value of *a* which is between 2 and 0. The value of $\vec{A}$ is also reduced by the value *a* and it is in the range of [-1,1] [2].

## Algorithm 1 Grey Wolf Optimization Algorithm

Initialize population *Xn ( n=1, 2, ....., m), Xα Xβ and Xδ*
Setting up *a* variable *and* both *A* and *C* vector
Calculating fitness value
    **While** (*i*< Maxi)
        **For** each search agent
            Update the location of the present search agent by Equation (5)
        **End For**
    Update *a, A,* and *C*
    Evaluating fitness
    Update *Xα, Xβ and Xδ*
    *i=i+1*
**End while**
return Xα

Besides the importance of optimization algorithms, data analysis is considered a great subject that can be worked on by researchers. Therefore, one of the well-known exploratory data analysis techniques is clustering that has been used to get an intuition about the structure of the data. K-means is one of the most popular unsupervised algorithms and the standard K-means algorithm was firstly presented by Stuart Lloyd in 1957 [68]. The algorithm can cluster the data into subgroups. Somehow, data points in each cluster are very similar compared to the data points in different clusters [70][51][71]. The algorithm follows a series of actions conducted to discover distinctive subgroups that are addressed below.

1. The main requirement of the K-means is defined by the number of subgroups as the K-means algorithm in data mining begins with the first group of randomly selected centroids for every cluster.
2. The next step is to take each data point belonging to a given data set and associate it to the nearest center by calculating the Euclidean distance from the centroid to all the data points. Then, calculating a new centroid for each cluster.
3. Keep doing the iteration if $K$ centers change their locations in the iteration until there is no change to the centroids.

By doing this producer, the algorithm tries to minimize the below objective function equation (8) as known as squared error function [72]:

$$J \sum_{i=1}^{c} \sum_{j=1}^{ci} \|\|x_i^j - cj\|\|\|^2 \tag{8}$$

Where,

$\|x_i^j - cj\|$ is the Euclidean distance between xi and cj.

$x_i^j$: is the number of data points in the ith cluster.

cj: is the number of cluster centers.

## 2.2 K-means Grey Wolf Optimization:

After initializing the population of GWO and its variables. K-means is used in the proposed KMGWO. K-means is used to divide the population of grey wolves into two clusters. Then, the fitness value is calculated for each cluster. After classifying the population into two clusters, a condition is added, which depends on a random number. The range of the random number is between 0 and1. Therefore, if the random value greater than 0.5 then KMGWO works on population clusters depending on their fitness. Inside the condition, the fitness value of both clusters is compared. If the fitness of cluster 1 is less than the fitness of cluster 2, then the position of the search agents is equalized to cluster position 1 and vise versa. However, if the random number is less than or equal to 0.5, KMGWO works on the basic population without any clustering. Therefore, this feature can be used with other algorithms but it requires evaluation to positively works well. However, in this paper, K-means are used as a feature to improve the performance of GWO. Moreover, the population is divided into 2 clusters because of achieving better results by KMGWO.

After selecting a specific cluster or working on the population without clustering, KMGWO attempts to calculate the objective function for each search agent until it finds the optimum fitness in the search space in the first iteration. Then, best search agents are calculated by using Equations (1), (2), (3), and (4). Next, updating each position is calculated by using Equation (5) and then these variables are updated for the next iteration, such as *a, A,* and *C*. Consequently, the best fitness in the iteration is selected and the positions are amended if they are out of the limitation. Finally, the best fitness is returned throughout the 500 iterations. Algorithm 2 shows the detail of KMGWO and Figure 1 illustrates the KMGWO diagram.

Algorithm 2 KMGWO

Initialize population *Xn ( n=1, 2, ....., m), Xα Xβ and Xδ*

Setting up *a* variable *and* both *A* and *C* vector
Use K-means to divide the population
Calculate *fitnessC1* and *fitnessC2*
    **If1** *rand > 0.5*
        **If2** *fitnessC1 < fitnessC2*
            *Postions=PositionsC1*
        **else**
            *Postions=PositionsC2*
        **EndIf2**
    **else**
        *Postions=Positions*
    **EndIf1**
    **While** (*i*< Maxi)
        **For** each search agent
            Update the location of the present search agent by Equation (5)
        **End For**
    Update *a, A*, and *C*
    Evaluating fitness
    Update $X_\alpha$, $X\beta$ *and* $X\delta$
    *i=i+1*
**End while**
return X$\alpha$

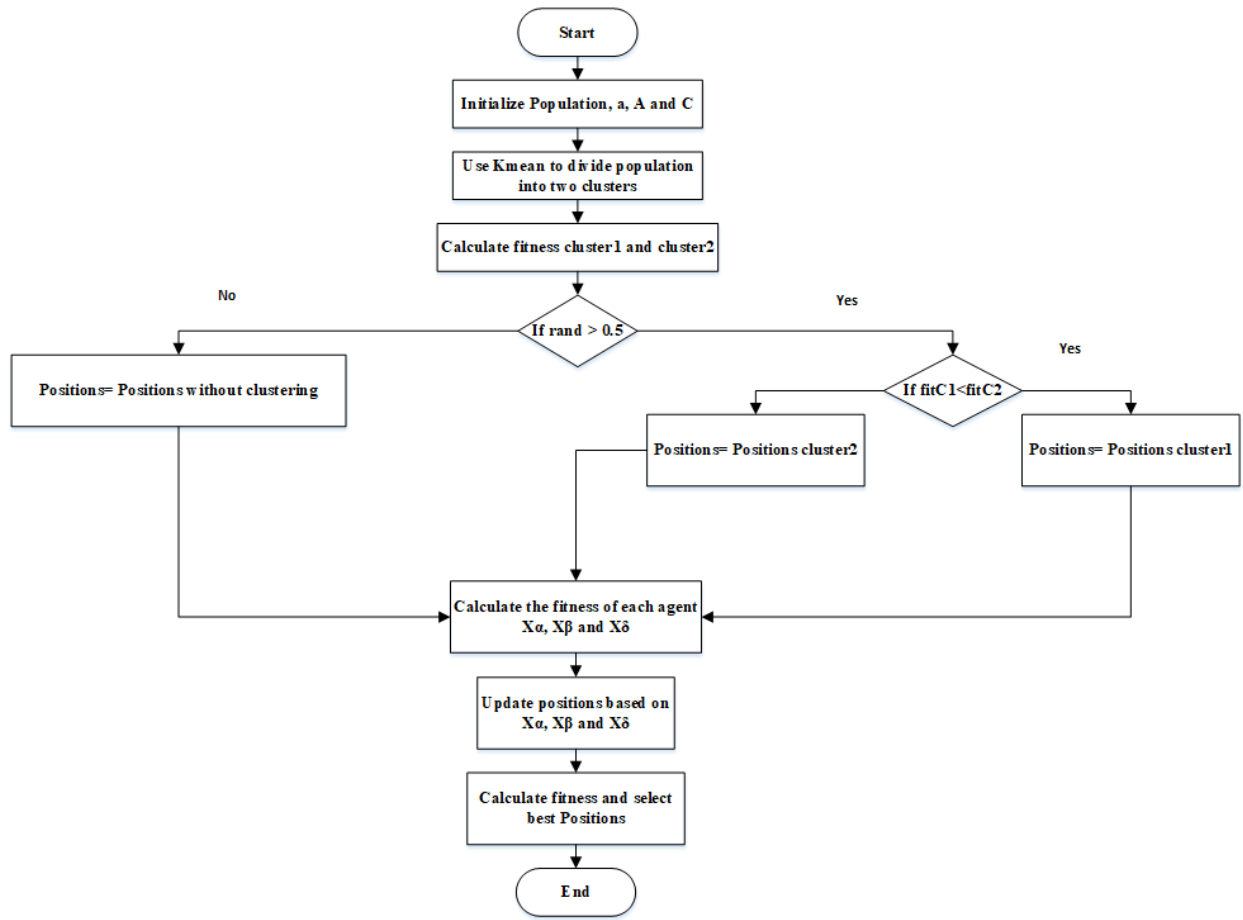

**Figure 1 KMGWO Flowchart**

# 3  Results and discussions

To ensure that K-means have an impact on other algorithms such as GWO, WOA, FDO, and WOA-BAT, we have applied K-means on these algorithms. Therefore, only GWO can achieve better results compared to other algorithms.

To evaluate KMGWO, the population of 30 agents is initialized with 500 iterations. KMGWO runs 30 times for each test function and then the average of the results is presented in Table 1.  CEC2019 benchmark functions consist of 10 functions, which are standard test functions, used to evaluate the capacity of metaheuristic algorithms. Therefore, GWO and KMGWO are also tested by using 10 benchmark functions.

Using the K-means technique, KMGWO improves the performance of GWO. Table 1 presents that KMGWO shows superior results in 6 functions out of 10. It also has the same results in 2 functions. Although, GWO is superior to KMWO in only 2 functions.

The reason behind this improvement is that K-means has a vital impact on the population of GWO and it divides the population into two clusters. Using K-means is useful for GWO because GWO traditionally works as three clusters as it has three wolves in the search space.

**Table 1 Comparison of GWO and KMGWO using CEC2019 Benchmark Test Functions.**

| F | GWO | | KMGWO | | Best |
|---|---|---|---|---|---|
| | Avg. | Std. | Avg. | Std. | |
| **1** | 2.13E+08 | 3.07E+08 | 4.06E+08 | 9.48E+08 | - |
| **2** | 18.3439 | 0.000304 | 18.344 | 0.000444 | + |
| **3** | 13.7024 | 7.23E-15 | 13.7025 | 0.000347 | + |
| **4** | 300.9815 | 686.8153 | **211.9193** | 489.3796 | + |
| **5** | 2.4313 | 0.251607 | **2.3949** | 0.271539 | + |
| **6** | 11.9394 | 0.730745 | 12.1168 | 0.690284 | - |
| **7** | 534.7655 | 292.0204 | **500.5104** | 322.4427 | + |
| **8** | 5.4021 | 0.993956 | **5.2551** | 0.837104 | + |
| **9** | 14.7328 | 49.95142 | **5.4874** | 1.584519 | + |
| **10** | 21.4973 | 0.068513 | **20.9642** | 2.986016 | + |

KMGWO also is compared to other common metaheuristic algorithms, such as Cat Swarm Optimizer (CSO), WOA-BAT [73], GWO, and WOA. As it is clear from Table 1, KMGWO is better than GWO. Therefore, it would be great if KMGWO is compared to other common algorithms. Consequently, Table 2 shows that KMGWO and WOA-BAT are competitive because KMGWO performs well in 4 functions out of 10. however, WOA-BAT has the second-best rank due to having the best results in 3 functions out of 10. The third best is CSO [74], which is superior in 2 functions while WOA is efficient in only 1 function. The worst algorithm is GWO which does not have effective results compared to the other four algorithms.

**Table 2 Comparison of CSO, WOA-BAT, GWO, WOA, and KMGWO using CEC2019 Benchmark Test Functions.**

| F | CSO | WOA-BAT | GWO | WOA | KMGWO | KMGWO |
|---|---|---|---|---|---|---|
| **1** | 1.58E+09 | **7.60E+07** | 2.13E+08 | 2.1E+10 | 4.06E+08 | - |
| **2** | 19.70367 | **17.5074** | 18.3439 | 18.355 | 18.344 | - |
| **3** | 13.70241 | **12.7029** | 13.7024 | 13.7024 | 13.7025 | - |
| **4** | **179.1984** | 2.12E+03 | 300.9815 | 347.6701 | 211.9193 | - |
| **5** | 2.671378 | 2.444 | 2.4313 | 3.0257 | **2.3949** | + |
| **6** | 11.21251 | 11.1491 | 11.9394 | **10.2783** | 12.1168 | - |
| **7** | **365.2358** | 606.299 | 534.7655 | 614.1979 | 500.5104 | - |
| **8** | 5.499615 | 5.7155 | 5.4021 | 6.027 | **5.2551** | + |
| **9** | 6.325862 | 22.816 | 14.7328 | 5.9336 | **5.4874** | + |
| **10** | 21.36829 | 21.1925 | 21.4973 | 21.2927 | **20.9642** | + |

To evaluate the average results that were achieved in Table 2, differences between each algorithm with KMGWO are calculated to find statistical results. Statistical results indicate that whether the results are significant or not. Table 3 shows that KMGWO has significant function results compared to CSO, WOA-BAT, GWO, and WOA. Overall, it can be said that KMGWO could achieve significant value against the above algorithms because of using clustering methods.

**Table 3 Statistical Results of KMGWO (*P*-value ) against  CSO, WOA-BAT, GWO, and WOA.**

| Differences | KMGWO vs. CSO | KMGWO vs. WOA-BAT | KMGWO vs. GWO | KMGWO vs. WOA |
|---|---|---|---|---|
| ***P-value*** | 0.000013 | 0.000013 | 0.000013 | 0.000009 |

# 4   Solving the Pressure Vessel Design Problem

Classical engineering problems, such as pressure vessel design is solved by different algorithms. Minimizing these sections which are forming, welding and material is the aim of this optimization. The head of the vessel has a hemispherical shape that is crapped at both ends [75], [76]. Four variables have to be minimized in this problem. Figure 2 shows the pressure vessel design variables. These four variables are:

- inner radius $R$,
- shell thickness $T_s$ ,
- head thickness $T_h$,
- cylindrical length section without counting the head $L$.

Consequently, four constraints of this problem have to be optimized. The constraints of the problem are shown below [77].

$$n = 1,2,3,4$$

$$\vec{x} = [x_1 x_2 x_3 x_4] = [T_s T_h\ R\ L],$$

$$f(\vec{x}) = 0.6224x_1x_3x_4 + 1.7781x_2x_3^2 + 3.1661x_1^2x_4 + 19.84x_1^2x_3, \qquad (9)$$

Variable limitation

$$0 \leq x_1 \leq 99,$$

$$0 \leq x_2 \leq 99,$$

$$10 \leq x_3 \leq 200,$$

$$10 \leq x_4 \leq 200,$$

These are subjected to

$$g_1(\vec{x}) = -x_1 + 0.0193x_3 \leq 0 \qquad (10)$$

$$g_2(\vec{x}) = -x_3 + 0.00954x_3 \leq 0 \qquad (11)$$

$$g_3(\vec{x}) = -\pi x_3^2 x_4 - \frac{4}{3}\pi x_3^3 + 1{,}296{,}000 \leq 0 \qquad (12)$$

$$g_4(\vec{x}) = x_4 + 240 \leq 0 \qquad (13)$$

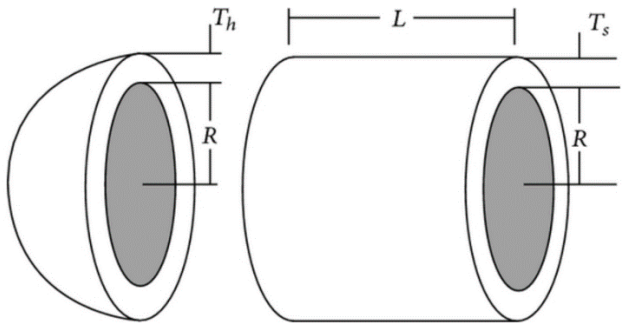

**Figure 2 Pressure Vessel Design Problem** [78]**.**

WOA, WOAGWO, and FDO had been used to solve this problem in the paper [30]. Therefore, FDO achieved better results compared to others. KMGWO is also used to solve this problem by using 30 populations with 500 iterations. Results illustrate that KMGWO enhances the performance of GWO and it is better than WOA, WOAGWO, and FDO. Therefore, using K-means as a clustering method enhances the performance of GWO (See Table 4).

**Table 4 Comparison WOA, WOAGWO, FDO, and KMGWO for Pressure Vessel Design**

| Run No. | WOA | WOAGWO | FDO | GWO | KMGWO |
|---|---|---|---|---|---|
| 1 | 13382.3 | 16697.97 | 377819.2 | 6916.404 | 6306.362 |
| 2 | 8106.258 | 13691.68 | 101800.8 | 6308.005 | 7030.402 |
| 3 | 13210.93 | 11391.75 | 57267.63 | 6325.155 | 6333.996 |
| 4 | 8369.843 | 11718.12 | 7556.644 | 6308.21 | 6348.065 |
| 5 | 8398.161 | 9536.122 | 6873.656 | 7260.878 | 6340.361 |
| 6 | 7438.549 | 11639.14 | 6777.363 | 7253.498 | 6297.992 |
| 7 | 19753.58 | 11127.94 | 19710.33 | 6310.451 | 6299.029 |
| 8 | 8205.785 | 15935.19 | 38812.92 | 6326.827 | 6317.929 |
| 9 | 8439.767 | 13165.6 | 8933.924 | 6346.056 | 6731.725 |
| 10 | 7966.724 | 10444.29 | 18258.82 | 6377.506 | 7332.575 |
| 11 | 8404.567 | 11303.79 | 116160.9 | 6299.859 | 6409.277 |
| 12 | 11396.46 | 16382.91 | 51808.24 | 6331.633 | 6606.493 |
| 13 | 10763.95 | 10050.45 | 83474.28 | 6950.996 | 6355.089 |
| 14 | 7714.56 | 11063.28 | 7538.922 | 7313.12 | 6697.121 |
| 15 | 159506.7 | 10520.38 | 47368.41 | 6431.194 | 6301.626 |
| Avg. | 20070.54 | 12311.24 | 63344.13 | 6603.986 | **6513.87** |
| Std. | 37401.74 | 2261.521 | 90960.63 | 391.9546 | 302.4419 |

# 5 Conclusions

The proposed algorithm is KMGWO which is based on GWO and K-means. GWO is an algorithm that is inspired by the attacking techniques of grey wolves. K-means technique is used as clustering with a condition to divide the population into two clusters to improve GWO in the search space. Results proved that KMGWO enhances the performance of GWO and obtained better results because of using a K-means algorithm. Therefore, it can be said that KMGWO improved the efficiency of GWO.

Therefore, to prove the improvement of KMGWO, KMGWO is tested using CEC2019 benchmark functions. Results proved that KMGWO is superior to GWO in 6 functions out of 10 while they are the same in only 2 functions. Then, differences between KMGWO and other algorithms are found to calculate the *p*-value. Results proved that KMGWO achieved the best significant value against CSO, WOA, and WOA-BAT.

To ensure that KMGWO would achieve high performance, classical engineering problems, such as pressure vessel design problems were taken into consideration. Results proved that KMGWO is very competitive and achieved optimum results compared to CSO, WOA-BAT, WOA, GWO.

Results proved that K-means has a great impact on the improvement of GWO. Therefore, this improvement was achieved because of the structure of the GWO algorithm since it works as clusters and it depends on three types of wolves. Consequently, similarities between K-means and GWO had improved and were used in KMGWO. KMGWO can be used to solve problems in the field of planning and medicine in the future.

## Acknowledgment

The authors wish to thank Charmo University, Sulaimani Polytechnic University, and the University of Kurdistan Hewler (UKH). This research did not receive any specific grant from funding agencies in the public, commercial, or not-for-profit sectors.

## Funding

## Conflicts of interest/Competing interests

There is no conflict of interest or competing interest is found.

## Availability of data and material

Data is available per request.

## Code availability

Data is available per request.